\documentclass[10pt,twocolumn,letterpaper]{article}

\usepackage{iccv}
\usepackage{times}
\usepackage{epsfig}
\usepackage{graphicx}
\usepackage{amsmath}
\usepackage{amsthm}
\usepackage{amssymb}
\usepackage{algorithm}
\usepackage{algpseudocode}
\usepackage{enumitem}

\graphicspath{ {./} }

\DeclareMathOperator*{\argmax}{argmax}

\usepackage[breaklinks=true,bookmarks=false]{hyperref}

\iccvfinalcopy 

\def\iccvPaperID{2400} 

\ificcvfinal\fi
\begin{document}

\title{Unsupervised object segmentation in video by \\ efficient selection of highly probable positive features}

\author{
  \begin{tabular}{c}
    Emanuela Haller$^{1,2}$\\
	{\tt\small haller.emanuela@gmail.com}
  \end{tabular}%
  \begin{tabular}{c}
    Marius Leordeanu$^{1,2}$\\
	{\tt\small marius.leordeanu@imar.ro}
  \end{tabular}\\ 
  \and
  \begin{tabular}{c}
    $^1$University Politehnica of Bucharest\\
    \small 313 Splaiul Independentei, Bucharest, Romania
  \end{tabular}%
  \begin{tabular}{c}
    $^2$Institute of Mathematics of the Romanian Academy\\
    \small 21 Calea Grivitei, Bucharest, Romania
  \end{tabular}\\ 
}

\maketitle

\begin{abstract}
We address an essential problem in computer vision, that of unsupervised object segmentation in video, where a main object of interest in a video sequence should be automatically separated from its background. 
An efficient solution to this task would enable large-scale video interpretation at a high semantic level in the absence of the costly manually labeled ground truth.
We propose an efficient unsupervised method for generating foreground object soft-segmentation masks based on automatic selection and learning from highly probable positive features.
We show that such features can be selected efficiently by taking into consideration the spatio-temporal, appearance and motion consistency of the object during the whole observed sequence. We also emphasize the role of the contrasting properties between the foreground object and its background. Our model is created in two stages: we start from pixel level analysis, on top of which we add a regression model trained on a descriptor that considers information over groups of pixels and is both discriminative and invariant to many changes that the object undergoes throughout the video.  
We also present theoretical properties of our unsupervised learning method, that under some mild constraints is guaranteed to learn a correct discriminative classifier even in the unsupervised case.
Our method achieves competitive and even state of the art results on the challenging Youtube-Objects and SegTrack datasets, while being at least one order of magnitude faster than the competition. We believe that the competitive performance of our method in practice, along with its theoretical properties, constitute an important step towards solving unsupervised discovery in video.
\end{abstract}

\section{Introduction}
    Unsupervised learning in video is a very challenging unsolved task in computer vision. Many researchers believe that solving this problem could shed new light on our understanding of intelligence and learning from a scientific perspective. It could also have a significant impact in many real-world computer and robotics applications. The task is drawing an increasing attention in the field due in part to the recent successes of deep neural network models in image recognition and to relatively low cost of collecting large datasets of unlabeled videos.

    There are several different published approaches for unsupervised learning and discovery of the main object of interest in video~\cite{sivic2005discovering, leordeanu2005unsupervised,papazoglou2013fast,li2013video}, but most have a high computational cost. In general, algorithms for unsupervised mining and clustering are expected to be computationally expensive due to the inherent combinatorial nature of the problem~\cite{jain1999data}.
        
    \begin{figure*}[!h]    
    \includegraphics[width=\textwidth]{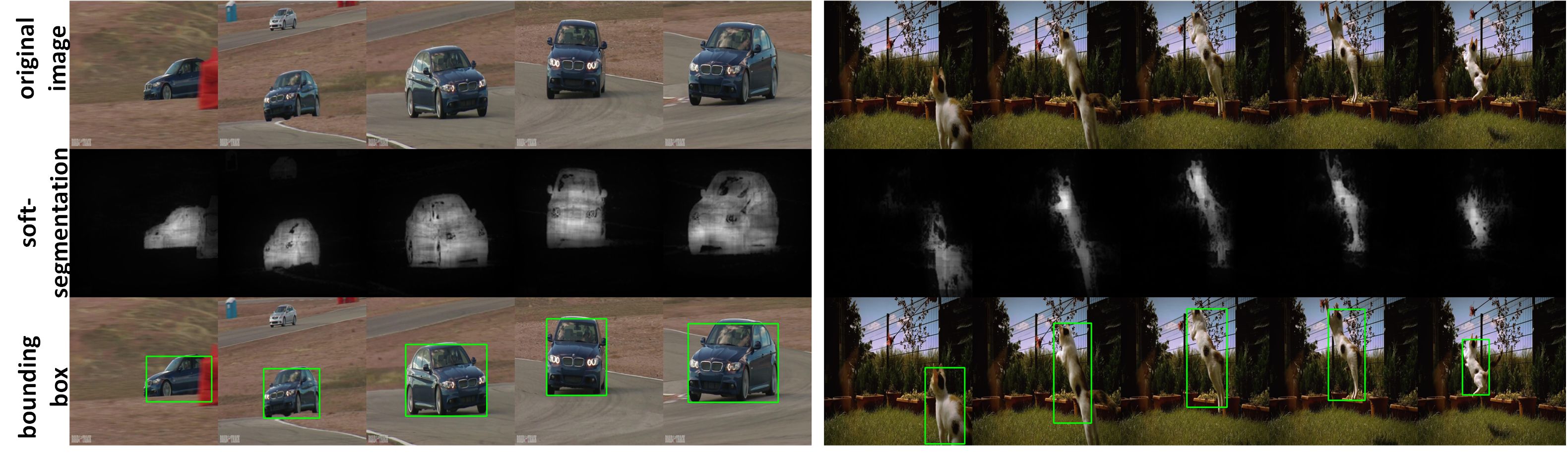}
    \centering
    \caption{Qualitative results of our method, which provides the soft-segmentation of the main object of interest and its bounding box.}
    \label{fig:algorithm_overview}
    \end{figure*}
        
    In this paper we address the computational cost challenge and propose a method that is both accurate and fast. We achieve our goal based on a key insight: we focus on
    selecting and learning from features that are highly correlated with the presence of the object of interest and can be rapidly selected and computed. \textbf{Note:} in this paper, when referring to highly probable positive features, we use "feature" to indicate a feature vector sample, not a feature type. While we do not require these features to cover all instances and parts of the object of interest (we could expect low recall), we show that is is possible to find, in the unsupervised case, positive features with high precision (a large number of those selected are indeed true positives). Then we prove theoretically that we can reliably train an object classifier using sets of positive and negative samples, both selected in an unsupervised way, as long as the set of features considered to be positive has high precision, regardless of the recall, if certain conditions are met (and they are often met in practice). We present an algorithm that can effectively and rapidly achieve this task in practice, in an unsupervised way, with state-of-the art results in difficult experiments, while being at least 10x faster than its competition. The proposed method outputs both the soft-segmentation of the main object of interest as well as its bounding box. Two examples are shown in Figure \ref{fig:algorithm_overview}. 
        
    While we do not make any assumption about the type of object present in the video, we do expect the sequence to contain a single main foreground object. The key insights that led to our formulation and algorithm are the following: 
    
    \noindent \textbf{1)} First, the foreground and background are complementary and in contrast to each other - they have different sizes, appearance and movements. We observed that the more we can take advantage of these contrasting properties the better the results, in practice. While the background occupies most of the image, the foreground is usually small and has distinct color and movement patterns - it stands out against its background scene. 
    
    \noindent \textbf{2)} The second main idea of our approach is that we should use this foreground-background complementarity in order to automatically select, with high precision, foreground features, even if the expected recall is low. Then, we could reliably use those samples as positives, and the rest as negative, to train a classifier for detecting the main object of interest. We present this formally in Sec \ref{subs:learning_HPP}. 
    
    These insights lead to our two main contributions in this paper: first, we show theoretically that by selecting features that are positive with highly probability, a robust classifier for foreground regions can be learned. Second, we present an efficient method based on this insight, which in practice outperforms its competition on many different object classes, while being 10x faster.
    
   \noindent \textbf{Related work on object discovery in video:}
   The task of object discovery in video has been tackled for many years, with early approaches being based on local features matching ~\cite{sivic2005discovering,leordeanu2005unsupervised}. Current literature offers a wide range of solutions, with varying degrees of supervision, going from fully unsupervised methods 
~\cite{papazoglou2013fast,li2013video} to partially supervised ones~\cite{jun2016pod,zhang2015semantic,zhang2013video,lee2011key,stretcu2015multiple} - which start from region, object or segmentation proposals estimated by systems trained in a supervised manner~\cite{alexe2012measuring,felzenszwalb2010object,endres2010category}. Some methods also require user input for the first frame of the video~\cite{jain2014supervoxel}. Most object discovery approaches that produce a fine shape segmentation of the object also make use of off-the-shelf shape segmentation methods~\cite{rother2004grabcut,fulkerson2009class,levinshtein2009turbopixels,carreira2012cpmc,li2013composite}. 
\section{Approach}
\begin{figure}[h]    
    \includegraphics[width=\linewidth]{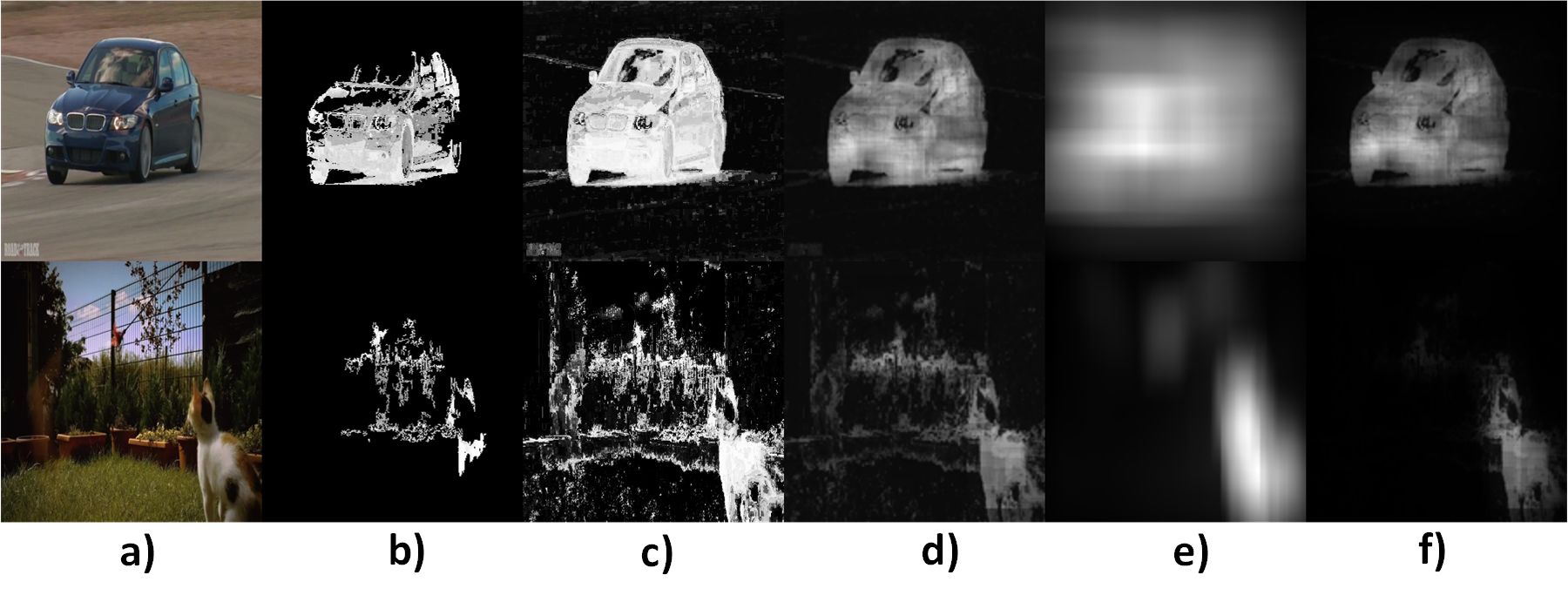}
    \centering
    \caption{Algorithm overview. 
    \textbf{a)} original image
    \textbf{b)} first pixel-level appearance model, based on initial object cues (Step 1 \& Step 2)
    \textbf{c)} refined pixel-level appearance model, built from the projection of soft-segmentation (Step 3 \& Step 4)
    \textbf{d)} patch-level appearance model (Step 5)
    \textbf{e)} motion estimation mask (part of Step 6)
    \textbf{f)} final soft-segmentation mask (Step 6).
    }
    \label{fig:mainFlow}
    \end{figure}

    Our method receives as input a video sequence, in which there is a main object of interest, and it outputs its soft-segmentation masks and associated bounding boxes. The proposed approach has, as starting point, a processing stage based on principal component analysis of the video frames, which provides an initial soft-segmentation of the object - similar to the recent VideoPCA algorithm introduced as part of the object discovery approach of ~\cite{stretcu2015multiple}. This soft-segmentation usually has high precision but may have low recall. Starting from this initial stage that classifies pixels independently based only on their individual color, next we learn a higher level descriptor that considers groups of pixel colors and is able to capture higher order statistics about the object properties, such as different color patterns and textures. During the last stage we combine the soft-segmentation based on appearance, with foreground cues computed from the contrasting motion of the main object vs. its scene. The resulting method is accurate and fast ($\approx 3$ fps in Matlab). Below, we summarize the steps of our approach (also see Figure \ref{fig:mainFlow}).  
    
    \begin{itemize}[noitemsep]
        \item[$\bullet$] Step 1: select highly probable foreground pixels based on the differences between the original frames and the frames projected on their subspace with principal component analysis (\ref{subs:videoPCA}).
        \item[$\bullet$] Step 2: estimate empirical color distributions for foreground and background from the pixel masks computed at Step 1. Use these distributions to estimate the probability of foreground for each pixel independently based on its color (\ref{subs:softSeg}).
        \item[$\bullet$] Step 3: improve the soft-segmentation from Step 2, by projection on the subspace of soft-segmentations (\ref{subs:softSegRec}).
        \item[$\bullet$] Step 4: re-estimate empirical color distributions for foreground and background from the pixel masks updated at Step 3. Use these distributions to estimate the probability of foreground for each pixel independently based on its color (\ref{subs:softSeg}).      
        \item[$\bullet$] Step 5: learn a discriminative classifier of foreground regions with regularized least squares regression on the soft segmentation real output $\in [0,1]$. Use a feature vector that considers groups of colors that co-occur in larger patches. Run classifier at each pixel location in the video and produce improved per frame foreground soft-segmentation (\ref{subs:classif}).
        \item[$\bullet$] Step 6: combine soft-segmentation using appearance (Step 5) with foreground motion cues efficiently computed by modeling the background motion. Obtain the final soft-segmentation (\ref{subs:motion}).
        \item[$\bullet$] Step 7: Optional: refine segmentation using GrabCut ~\cite{rother2004grabcut}, by considering as potential foreground and background samples the pixels given by the soft-segmentation from Step 6. (\ref{subs:refinement})
    \end{itemize}
    
	We reiterate: our algorithm has at its core two main ideas. The first is that the object and the background have contrasting properties in terms of size, appearance and movement. This insight leads to the ability to select reliably a few regions in the video that are highly likely to belong to the object. The following, second idea, which brings certain formal guarantees, is that if we are able to select in an unsupervised manner, even a small portion of the foreground object but with high precision then, under some reasonable assumptions, we could train a robust foreground-background classifier that can be used for the automatic discovery of the object. The pseudo-code of our approach is presented in Algorithm \ref{alg:fullAlgorithm}. In Table \ref{tab:stagesEval} we introduce quantitative results of the different stages of our Algorithm, along with the associated execution times. In Table \ref{tab:Evolution} we present the improvements in precision, recall and f-measure between the different steps of our algorithm. Note that the arrows go from the precision and recall of 
	the samples considered to be positive during training, to the precision and recall of the pixels classified during testing. The significant improvement in f-measure is explained by our theoretical results (stated in Proposition 1), which shows that under certain conditions, a reliable classifier will be learned even if the recall of the corrupted positive samples is low, as long as the precision is relatively high.
		
	\begin{algorithm}
        \caption{Video object segmentation}
        \label{alg:fullAlgorithm}
        \begin{algorithmic}[1]
            \State get input frames $\mathbf{F}^i$
            \State PCA($\mathbf{A}_1$) $=>\mathbf{V}_1$ eigenvectors; $\mathbf{A}_1(i,:)=\mathbf{F}^i(:)$
            \State $\mathbf{R}_1=\mathbf{\bar A}_1+(\mathbf{A}_1-\mathbf{\bar A}_1)*\mathbf{V}_1*\mathbf{V}_1^T$ - reconstruction
            \State ${\mathbf{P}_1}^i=\mathbf{d}(\mathbf{A}_1(i,:),\mathbf{R}_1(i,:))$
            \State ${\mathbf{P}_1}^i={\mathbf{P}_1}^i\otimes \mathbf{G}_{{\sigma}_1}$ 
            \State ${\mathbf{P}_1}^i=>$ pixel-level appearance model $=>{\mathbf{S}_1}^i$ 
            \State PCA($\mathbf{A}_2$) $=>\mathbf{V}_2$ eigenvectors; $\mathbf{A}_2(i,:)={\mathbf{S}_1}^i(:)$
            \State $\mathbf{R}_2=\mathbf{\bar A}_2+(\mathbf{A}_2-\mathbf{\bar A}_2)*\mathbf{V}_2*\mathbf{V}_2^T$ - reconstruction
            \State ${\mathbf{P}_2}^i={\mathbf{R}_2}^i\otimes \mathbf{G}_{{\sigma}_2}$ 
            \State ${\mathbf{P}_2}^i=>$ pixel-level appearance model $=>{\mathbf{S}_2}^i$ 
            \State $\mathbf{D}$ - data matrix containing patch-level descriptors
            \State $\mathbf{s}$ patch labels extracted from ${\mathbf{S}_2}^i$
            \State select $k$ features from $\mathbf{D}$ $=>\mathbf{D_s}$
            \State $\mathbf{w}={(\lambda \mathbf{I}+{\mathbf{D}_s}^T\mathbf{D}_s)}^{-1}{\mathbf{D}_s}^T\mathbf{s}$
            \State evaluate $=>$  patch-level appearance model $=>{\mathbf{S}_3}^i$
            \For{each frame $i$}
                \State compute $\mathbf{I}_x$, $\mathbf{I}_y$ and $\mathbf{I}_t$
                \State build motion matrix $\mathbf{D}_m$
                \State $\mathbf{w}_m={({\mathbf{D_m}}^T\mathbf{D_m})}^{-1}           {\mathbf{D}_m}^T\mathbf{I}_t$
                \State compute motion model $\mathbf{M}^i$
                \State ${\mathbf{M}}^i={\mathbf{M}}^i\otimes {\mathbf{G}_{\sigma}}^i$ 
                \State combine ${\mathbf{S}_3}^i$ and $\mathbf{M}^i$ $=>{\mathbf{S}_4}^i$
            \EndFor
        \end{algorithmic}
        \end{algorithm}
	
	    \subsection{Select highly probable object regions} \label{subs:videoPCA}
		We estimate the initial foreground regions by Principal Component Analysis, an approach similar to the recent method for soft foreground segmentation, VideoPCA~\cite{stretcu2015multiple}. Other approaches for soft foreground discovery could have been applied here, such as~\cite{zitnick2014edge, hou2007saliency, jiang2013salient}, but we have found the direction using PCA to be both fast and reliable and to fit perfectly with the later stages of our method. At this step we first project the frames on their subspace using PCA and compute reconstruction error images as differences between the original frames and their PCA reconstructed counter parts.  If principal components are $\mathbf{u}_i, \; i \in [0 \dots n_u]$ (we used $n_u = 3$) and frame $\mathbf{f}$ projected on the subspace is $\mathbf{f}_r \approx \mathbf{f}_0 + \sum_{i=1}^{n_u} ((\mathbf{f}-\mathbf{f}_0)^\top\mathbf{u}_i)\mathbf{u}_i$, then we compute the error image $\mathbf{f}_{diff} = |\mathbf{f} -  \mathbf{f}_r|$. High value pixels in the error image are more likely to belong to foreground. If we further smooth these regions with a large enough Gaussian and multiply the resulting smoothed difference with another large centered Gaussian (which favors objects in the center of the image), we obtain soft foreground masks that have high precision (most pixels on these masks indeed belong to true foreground), even though they often have low recall (only a small fraction of all object pixels are selected). As discussed, high precision and low recall is all need at this stage (see Table \ref{tab:Evolution})
		
	    \subsubsection{Initial soft-segmentation} \label{subs:softSeg}
	    Considering the small fraction of the object regions obtained at the previous step, the initial whole object soft segmentation is computed by capturing foreground and background color distributions, followed by an independent pixel-wise classification. Let $p(c|fg)$ and $p(c|bg)$ be the true foreground ($fg$) and background ($bg$) probabilities for a given color $c$. Using Bayes' formula with equal priors, we compute the probability of foreground for a given pixel, with an associated color $c$, as $p(fg|c)=\frac{p(c|fg)}{p(c|fg)+p(c|bg)}$. 
	    The foreground color likelihood is computed as $p(c|fg)=\frac{n(c,fg)}{n(c)}$, where $n(c,fg)$ is the number of considered foreground pixels having color $c$ and $n(c)$ is the total number of pixels having color $c$. The background color likelihood is computed in a similar manner. Note that when computing the color likelihoods, we take into consideration information gathered from the whole movie, obtaining a robust model. The initial soft segmentation produced here is not optimal but it is computed fast ($50-100$ fps) and of sufficient quality to insure the good performance of the subsequent stages. The first two steps of the method follow the algorithm VideoPCA first proposed in~\cite{stretcu2015multiple}. 
	    In the next Sec.~\ref{subs:learning_HPP} we present and prove our main theoretical result (Proposition 1), which explains in large part why our approach is able to produce accurate object segmentations in an unsupervised way.
	    
	    \subsection{Learning with HPP features}
	    \label{subs:learning_HPP}
	    
	    \begin{figure}[h]    
        \includegraphics[width=\linewidth]{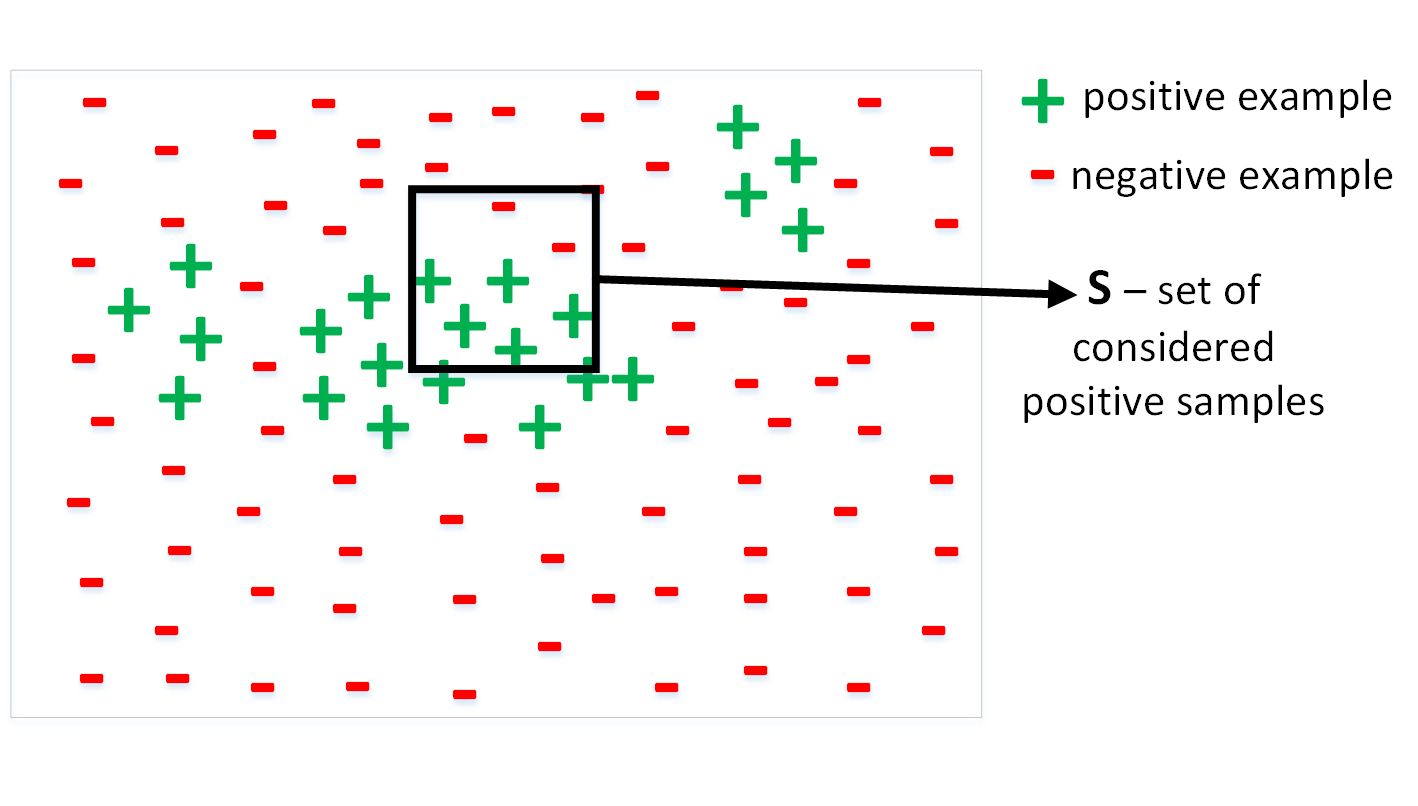}
        \centering
        \caption{Learning with HPP feature vectors. Essentially, Proposition 1 shows that we could learn a reliable discriminative classifier from a small set of corrupted positive samples, with the rest being considered negative, if the corrupted positive set contains mostly good features such that the ratio of true positives in the corrupted positive set is greater than the overall ratio of true positives. This assumption can often be met in practice and efficiently used for unsupervised learning.}
        \label{fig:posNegSel}
        \end{figure}
        
	    In Proposition 1 we show that a classifier trained on corrupted sets of positive and negative samples, can learn the right thing as if true positives and negatives were used for training, if the following condition is met: the set of corrupted positives should contain positive samples in a proportion that is greater than the overall proportion of true positives in the whole training set. Let us start with the example in Figure \ref{fig:posNegSel}, where we have selected a set of samples $S$, inside the box, as being positive. The set $S$ has high precision (most samples are indeed positive), but low recall (most true positives are wrongly labeled). Next we show that the sets $S$ and $\neg S$ could be used reliably (as defined in Proposition 1, below) to train a binary classifier.
	    	    
	    Let $p(E_+)$ and $p(E_-)$ be the true distributions of positive and negative elements, and $p(\mathbf{x}|S)$ and $p(\mathbf{x}|\neg S)$ be the probabilities of observing a sample inside and outside the considered positive set $S$ and negative set $\neg S$, respectively. 
	    
	    \textbf{Proposition 1} (learning from highly probable positive (HPP) features):
	        Considering the following hypotheses $\mathbf{H_1}:p(E_+)<q<p(E_-)$, $\mathbf{H_2}:p(E_+|S)>q>p(E_-|S)$, where $q \in (0,1)$, and $\mathbf{H_3}:p(\mathbf{x}|E_+) \text{ and } p(\mathbf{x}|E_-)$ are independent of $S$, then, for any sample $\bf x$ we have: 
	        $p(\mathbf{x}|S)>p(\mathbf{x}|\neg S) <=> p(\mathbf{x}|E_+)>p(\mathbf{x}|E_-)$. In other words, a classifier that classifies pixels based on their likelihoods w.r.t to $S$ and $\neg S$ will take the same decision as if it was trained on the true positives and negatives, and we refer to it as a \emph{reliable} classifier.
	        
	    \textbf{Proof:} We express $p(E_-)$ as $\frac{(p(E_-)-p(E_-|S)\cdot p(S))}{(1-p(S))}$ (Eq 1), using the hypothesis and the sum rule of probabilities. Considering (Eq 1), hypothesis $\mathbf{H_1}$, $\mathbf{H_2}$, and the fact that $p(S)>0$, we obtain that  $p(E_-|\neg S)>q$ (Eq 2). In a similar fashion, $p(E_+|\neg S)<q$ (Eq 3). The previously inferred relations (Eq 2 and Eq 3) generate $p(E_-|\neg S)>q>p(E_+|\neg S)$ (Eq 4), which along with hypothesis $\mathbf{H_2}$ help as conclude that 
	    $p(E_+|S)>p(E_+|\neg S)$ (Eq 5). 
	    Also, from $\mathbf{H_3}$, we infer that $p(\mathbf{x}|E_+,S)=p(\mathbf{x}|E_+)$ and  $p(\mathbf{x}|E_-,S)=p(\mathbf{x}|E_-)$ (Eq 6).
	    Using the sum rule and hypothesis $\mathbf{H_3}$, we obtain that $p(\mathbf{x}|S)=p(E_+|S)\cdot (p(\mathbf{x}|E_+)-p(\mathbf{x}|E_-))+p(\mathbf{x}|E_-)$ (Eq 7). In a similar way, it results that $p(\mathbf{x}|\neg S)=p(E_+|\neg S)\cdot (p(\mathbf{x}|E_+)-p(\mathbf{x}|E_-))+p(\mathbf{x}|E_-)$ (Eq 8).
	   
	    $p(\mathbf{x}|S)>p(\mathbf{x}|\neg S) => p(\mathbf{x}|E_+)>p(\mathbf{x}|E_-)$: using the hypothesis and previously inferred results (Eq 5, 7 and 8) it results that $p(\mathbf{x}|E_+)>p(\mathbf{x}|E_-)$.
	    
	    $p(\mathbf{x}|E_+)>p(\mathbf{x}|E_-) => p(\mathbf{x}|S)>p(\mathbf{x}|\neg S)$: from the hypothesis we can infer that $p(\mathbf{x}|E_+)-p(\mathbf{x}|E_-)>0$, and using (Eq 5) we obtain $p(\mathbf{x}|S)>p(\mathbf{x}|\neg S)$.   
	    $\Box$
	    
	    \begin{table}
	    \begin{center}
	    \begin{tabular}{|l|c||c||c|}
	    \hline 
	                     & \shortstack[c]{Step 1 \\ Step 2} 
	                     & \shortstack[c]{Step 3 \\ Step 4} 
	                     & Step 5 \\
	    \hline\hline
	    precision       & $66 \rightarrow 70$ & $62 \rightarrow 60$ & $64 \rightarrow 74$ \\
	    recall          & $17 \rightarrow 51$ & $45 \rightarrow 60$ & $58 \rightarrow 68$ \\
	    \hline
	    f-measure       & $27 \rightarrow 59$ & $53 \rightarrow 60$ & $61 \rightarrow 72$ \\
	    \hline
	    \end{tabular}
	    \end{center}
	    \caption{Evolution of precision, recall and f-measure of the feature samples considered as positives (foreground) at different stages of our method. We start with a corrupted set of positive samples with high precision and low recall, and improve both precision and recall through the stages of our method. Thus the soft masks become more and more accurate from one stage to the next.} \label{tab:Evolution}
	    \end{table}
	   
	\subsection{Object proposals refinement} \label{subs:softSegRec}
	    The result presented above provides a basis for both stages of our method, the one that classifies pixels independently based on their colors, and the second in which we consider higher order color statistics among groups of pixels that belong to different image patches. First, we improve the soft segmentations obtained so far, by projecting them on their PCA subspace. Instead of looking at the differences between original input and its PCA reconstruction, we now simply replace the soft segmentations with the PCA projected versions (using 8 principal components), thus reducing the amount of noise that might be leftover from the previous steps. The pseudocode of this step can also be seen in Algorithm~\ref{alg:fullAlgorithm}. 
	    	    
	\subsection{Considering color co-occurrences} \label{subs:classif}
	
\begin{figure}[h]    
        \includegraphics[width=0.7\linewidth]{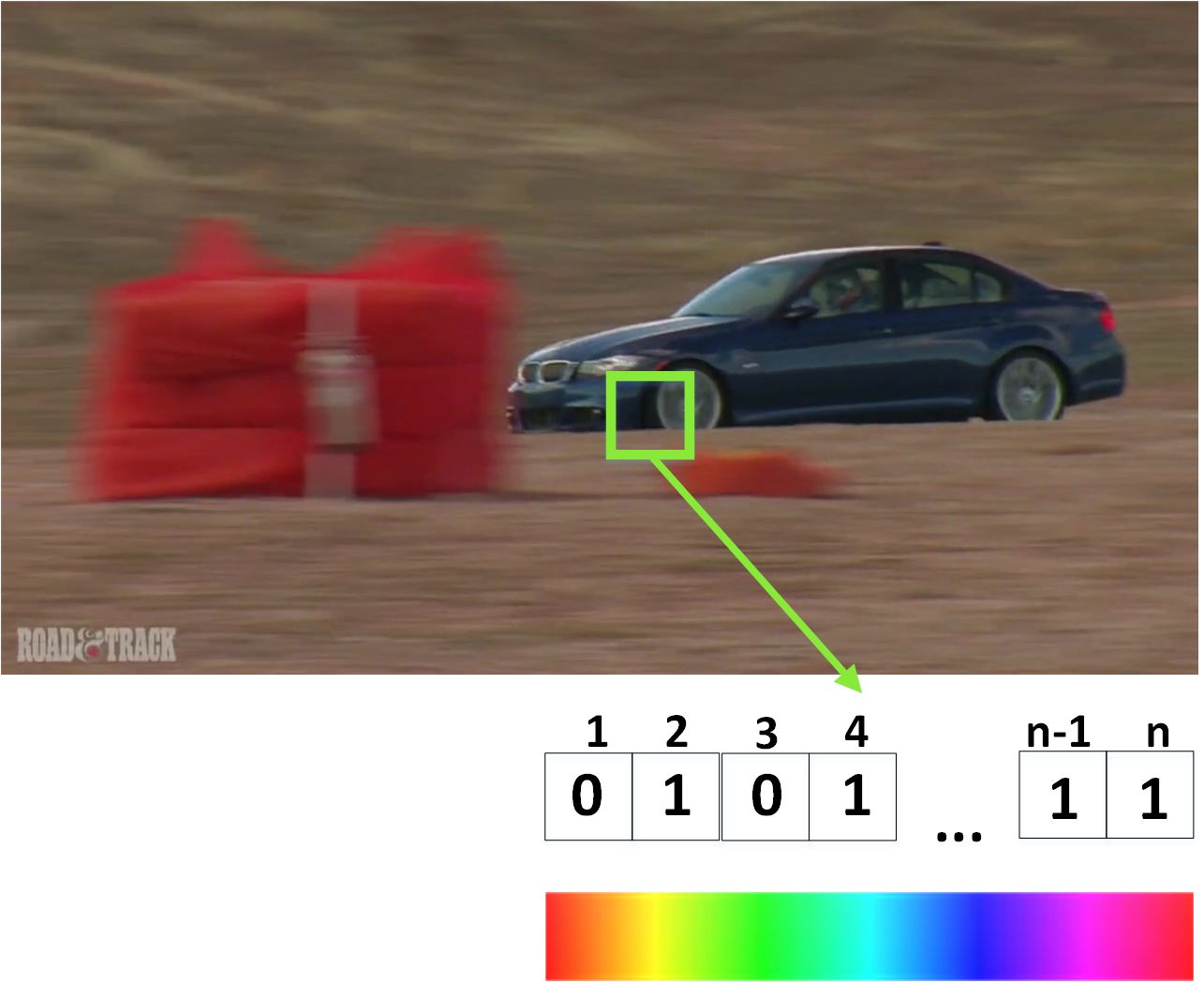}
        \centering
        \caption{Initial patch descriptors encoding color occurrences ($n$ number of considered colors).}
        \label{fig:patchBasedReg}
        \end{figure}

	    The foreground masks obtained so far were computed by treating each pixel independently, which results in masks that are not always correct, as first-order statistics, such as colors of individual pixels, cannot capture more global characteristics about object texture and shape. At this step we move to the next level of abstraction by considering groups of colors present in local patches, which are sufficiently large to capture object texture and local shape.
	    We define a patch descriptor based on local color occurrences, as 
	    an indicator vector $\mathbf{d}_W$ over a given patch  window $W$, such that $\mathbf{d}_W(c)=1$ if color $c$ is present in window $W$ and $0$ otherwise (Figure \ref{fig:patchBasedReg}). Colors are indexed according to their values in HSV space, where channels H, S and V are discretized in ranges $[1,15]$, $[1,11]$ and $[1,7]$, generating a total of $1155$ possible colors. The descriptor does not take in consideration the exact spatial  location of a given color in the patch, nor its frequency. It only accounts for the presence of $c$ in the patch. This leads to invariance to most rigid or non-rigid transformations, while preserving the local appearance characteristics of the object. Then, we take a classification approach and learn a classifier (using regularized least squares regression, due to its considerable speed and efficiency) to separate between highly probable positive (HPP) descriptors and the rest, collected from the whole video according to the soft masks computed at the previous step. The classifier is generally robust to changes in viewpoint, scale, illumination,
	    and other noises, while remaining discriminative (Figure \ref{fig:mainFlow}).
	       
	    \noindent \textbf{Unsupervised descriptor learning:} 
	    Not all $1155$ colors are relevant for our classification problem. Most object textures are composed of only a few important colors that distinguishes them against the background scene. Effectively reducing the number of colors in the descriptor and selecting only the relevant ones can improve both speed and performance. We use the efficient selection algorithm presented in~\cite{leordeanu2015labeling}. The method proceeds as follows.  Let  $n$ be the total number of colors and $k<n$ the number of relevant colors we want to select. The idea is to identify the group of $k$ colors with the largest amount of covariance - they will be the ones most likely to select well the foreground versus the background (see ~\cite{leordeanu2015labeling} for details). Now consider $\mathbf{C}$ the covariance matrix of the colors forming the rows in the data matrix $\mathbf{D}$. The task is to solve the following optimization problem:
	    \begin{equation} \label{eq:featSel}
	        \begin{split}
	            \mathbf{w}^*=\argmax_\mathbf{w} \mathbf{w}^T\mathbf{C}\mathbf{w} \\
	            s.t. \sum_{i=1}^{n} w_i=1, w_i\in [0,\frac{1}{k}]
	        \end{split}
	    \end{equation}
	    
	    The non-zero elements of $\mathbf{w}^*$ correspond to the colors we need to select for creating our descriptor used by the classifier (based on regularized least squares regression model), so we define a binary mask $\mathbf{w}_s\in \mathbb{R}^{n\text{x}1}$ over the colors (that is, the descriptor vector) as follows:
	    \begin{equation} \label{featBM}
            \mathbf{w_s}(i) = 
            \begin{cases}
                1 & \text{if } \mathbf{w}^*(i) > 0\\
                0 & \text{otherwise}
            \end{cases}
        \end{equation}
        The problem above is NP-hard, but a good approximation can be efficiently found by the method presented in~\cite{leordeanu2015labeling}, based on a convergent series of integer projections on the space of valid solutions. 
        
\begin{figure}[h]    
        \includegraphics[width=\linewidth]{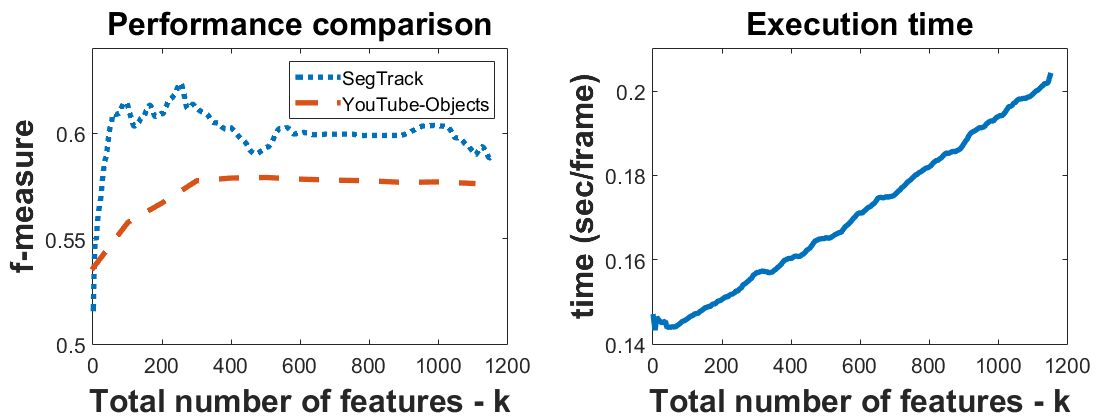}
        \centering
        \caption{Features selection - optimization and sensitivity analysis.}
        \label{fig:kTimePlot}
        \end{figure}

	    Next we define $\mathbf{D}_s\in \mathbb{R}^{m\text{x}(1+k)}$ to be the data matrix, with a training sample per row, after applying the selection mask to the descriptor; $m$ is the number of training samples and $k$ is the number of colors selected to form the descriptor; we add a constant column of 1's for the bias term.
	    Then the weights $\mathbf{w}  \in \mathbb{R}^{(1+k)\text{x}1}$ of the regularized regression model are learned very fast, in closed-form:
	    \begin{equation} \label{regModel}
	            \mathbf{w}={(\lambda \mathbf{I}+{\mathbf{D}_s}^T\mathbf{D}_s)}^{-1}{\mathbf{D}_s}^T\mathbf{s}
	    \end{equation}
	    where $\mathbf{I}$ is the identity matrix, $\lambda$ is the regularization term  and $\mathbf{s}$ is the vector of soft-segmentation masks values (estimated at the previous step) corresponding to the samples chosen for training of the descriptor. 
	    The optimal number of selected colors is a relatively small fraction of the total number, as expected. Besides the slight increase in performance, the real gain is in the significant decrease in computation time (see Figure \ref{fig:kTimePlot}). Then, the final appearance based 
	    soft-segmentation masks are generated by evaluating the regression model for each pixel. 
	    
	\subsection{Combining appearance and motion}  
	\label{subs:motion}
		The foreground and background have complementary properties at many levels, not just that of appearance. Here we consider that the object of interest must distinguish itself from the rest of the scene in terms of its motion pattern. A foreground object that does not move in the image, relative to its background, cannot be discovered using information from the current video alone. We take advantage of this idea by the following efficient approach.
			
		Let $\mathbf{I}_t$ be the temporal derivative of the image as a function of time, estimated as difference between subsequent frames $\mathbf{I}_{t+1}-\mathbf{I}_{t}$. Also let $\mathbf{I}_x$ and $\mathbf{I}_y$ be the partial derivatives in the image w.r.t $x$ and $y$. 
		Consider $\mathbf{D}_m$ to be the motion data matrix, with one row  per pixel $p$ in the current frame corresponding to   $[\mathbf{I}_x, \mathbf{I}_y, x\mathbf{I}_x, x\mathbf{I}_y, x\mathbf{I}_y, y\mathbf{I}_y]$ at locations estimated as background by the foreground segmentation estimated so far. Given such a matrix at time $t$ we linearly regress $\mathbf{I}_t$ on $\mathbf{D}_m$.  The solution would be a least square estimate of an affine motion model for the background using first order Taylor expansion of the image w.r.t time: $\mathbf{w}_m={({\mathbf{D}_m}^T\mathbf{D}_m)}^{-1}{\mathbf{D}_m}^T\mathbf{I}_t$.
	    Here $\mathbf{w_m}$ contains the six parameters defining the affine motion (including translation) in 2D.	
		
	    Then, we consider deviations from this model as potential good candidates for the presence of the foreground object, which is expected to move differently than the background scene. The idea is based on an approximation, of course, but it is very fast to compute and can be reliably combined with the appearance soft-segmentation mask. Thus we evaluate the model in each location $p$ and compute errors $|\mathbf{D}_m(p)\mathbf{w}_m - \mathbf{I}_t(p)|$. We normalize the error image and map  it to $[0,1]$. This produces a soft mask (using motion only) of locations that do not obey the motion model - they are usually correlated with object locations. This map is then smoothed with a Gaussian (with $\sigma$ proportional to the distribution on $x$ and $y$ of the estimated object region).
	    
	    At this point we have a soft object segmentation computed from appearance alone, and a second one computed, independently based on motion cues. We then multiply the soft results of the two independent pathways and obtain the final segmentation.
	       
	    \paragraph{Optional: refinement of video object segmentation} \label{subs:refinement}
	    Optionally we can further refine the soft-mask by applying an off-the-shelf segmentation algorithm, such as GrabCut ~\cite{rother2004grabcut} and feeding it our soft foreground segmentation. \textbf{Note:} in our experiments we used GrabCut only for evaluation on SegTrack, where we were interested in the fine details of the objects shape. All other experiments are performed without this step.
	    
\section{Experimental analysis}
\begin{figure*}[t]    
    \includegraphics[width=\textwidth]{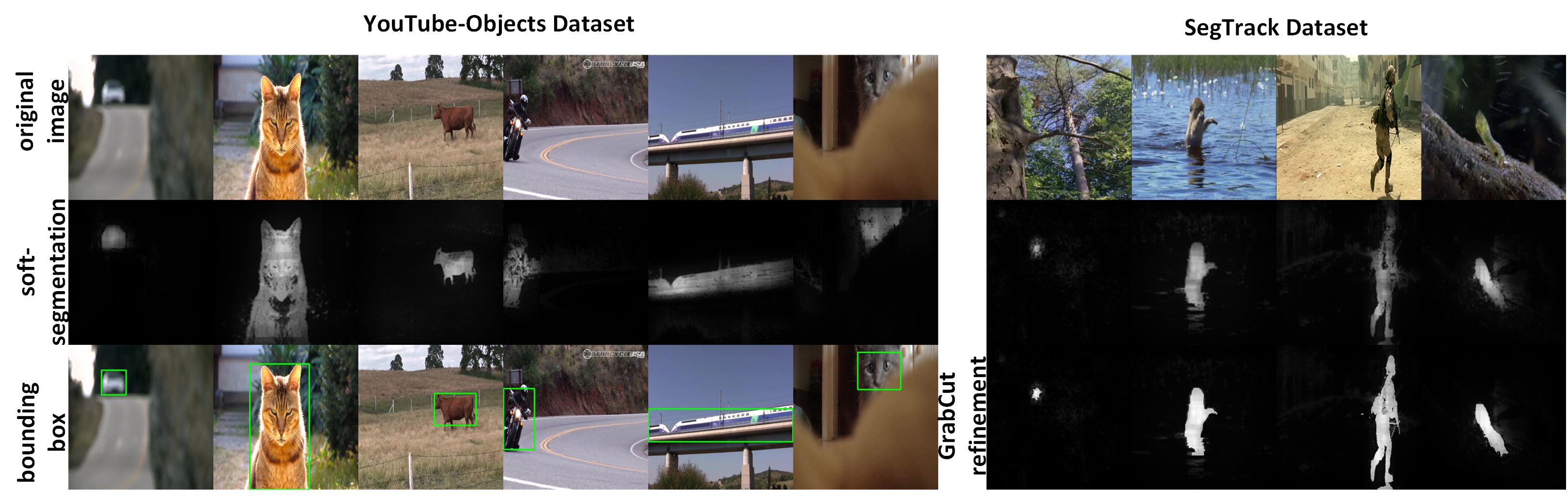}
    \centering
    \caption{Qualitative results on YouTube-Objects dataset and SegTrack dataset.}
    \label{fig:example}
    \end{figure*}

    Our experiments were performed on two datasets: YouTube-Objects dataset and SegTrack v2 dataset. We first introduce some qualitative results of our method, on the considered datasets (Figure \ref{fig:example}). Note that for the final evaluation on the YouTube-Objects dataset, we also extract object bounding boxes, that are computed using the distribution of the pixels with high probability of being part of the foreground. Both position and size of the boxes are computed using a mean shift approach. 
    For the final evaluation on the SegTrack dataset, we have refined the soft-segmentation masks, using the GrabCut algorithm ~\cite{rother2004grabcut}.
    In Tabel \ref{tab:stagesEval} we present evaluation results for different stages of our algorithm, along with the execution time, per stage. The f-measure is increased with each stage of our algorithm.  
    
        \begin{table}
	    \begin{center}
	     \resizebox{0.8\columnwidth}{!}{%
	    \begin{tabular}{|l|c|c|c|c|}
	    \hline 
	                                                & Step 2      & Step 4  & Step 5    & Step 6   \\
	    \hline\hline
	    f-meas. (SegTrack)        & 59.0 & 60.0 & 72.0 & 74.6 \\
	    \hline
	    f-meas. (YTO)             & 53.6 & 54.5 & 58.8 & 63.4 \\
	    \hline\hline    
	    Runtime     & 0.05 & 0.03 & 0.25 & 0.02 \\
	    \hline
	    \end{tabular}
	    }
	    \end{center}
	    \caption{Performance analysis and execution time for all stages of our method.} \label{tab:stagesEval}
	    \end{table}
			
	\subsection{YouTube-Objects dataset}
	    \textbf{Dataset.}
	    The YouTube-Objects dataset ~\cite{prest2012learning} contains a large number of videos filmed in the wild, collected from YouTube. It contains challenging, unconstrained sequences of ten object categories (aeroplane, bird, boat, car, cat, cow, dog, horse, motorbike, train). The sequences are considered to be challenging as they are completely unconstrained, displaying objects performing rapid movements, with difficult dynamic backgrounds, illumination changes, camera motion, scale and viewpoint changes and even editing effects, like flying logos or joining of different shots. The ground truth is provided for a small number of frames, and contains bounding boxes for the object instances. Usually, a frame contains only one primary object of the considered class, but there are some frames containing multiple instances of the same class of objects. Two versions of the dataset were released, the first (YouTube-Objects v1.0) containing 1407 annotated objects from a total of ~570 000 frames, while the second (YouTube-Objects v2.2) contains 6975 annotated objects from ~720 000 frames.
	    	    
	    \textbf{Metric.}
	    For the evaluation on the YouTube-Objects dataset we have adopted the CorLoc metric, computing the percentage of correctly localized object bounding-boxes. We evaluate the correctness of a provided box using the PASCAL-criterion (intersection over union $\ge 0.5$).
	    
	    \textbf{Results.} We compare our method against ~\cite{jun2016pod, zhang2015semantic, prest2012learning, stretcu2015multiple, papazoglou2013fast}. We considered their results as originally reported in the corresponding papers. The comparison is presented in Table \ref{tab:YTO}. From our knowledge, the other methods were evaluated on YouTube-Objects v1.0, on the training samples (the only exception would be ~\cite{stretcu2015multiple}, where they have considered the full v1.0 dataset). Considering this, and the differences between the two versions, regarding the number of annotations, we have reported our performances on both versions, in order to provide a fair comparison and also to report the results on the latest version, YouTube-Objects v2.2 (not considered for comparison). We report results of the evaluation on v1.0 by only considering  the training samples, for a fair comparison with other methods. Our method, which is unsupervised, is compared against both supervised and unsupervised methods. In the table, we have marked state-of-the-art results for unsupervised methods (bold), and overall state-of-the-art results (underlined). We also mention the execution time for the considered methods, in order to prove that our method is one order of magnitude faster than others (see section \ref{subs:compTime} for details).  	    
	    
	    The performances of our method are competitive, obtaining state-of-the-art results for 3 classes, against both supervised and unsupervised methods. If we report only to the unsupervised methods, we obtain state-of-the-art results for 7 classes. On average, our method performs better than all the others, and also in terms of execution time (also see section \ref{subs:compTime}). The fact that, on average, our algorithm outperforms other methods proves that it generalizes better for different classes of objects and different types of videos. Our solution performs poorly on the "horse" class, as many sequences contain multiple horses, and our method is not able to correctly separate the instances. Another class with low performances is the "cow" class, where we deal with same problems as in the case of "horse" class, and in addition, the objects are usually still, being hard to segment in our system.

        \begin{table}[h]
	    \begin{center}
	    \resizebox{\columnwidth}{!}{%
	    \begin{tabular}{|l|c|c||c|c|c|c||c|}
	    \hline 
	    
	    \shortstack{Method \\ Supervised?}	    
	    & \shortstack{~\cite{jun2016pod} \\ Y} 
	    & \shortstack{~\cite{zhang2015semantic} \\ Y} 
	    & \shortstack{~\cite{prest2012learning} \\ N}
	    & \shortstack{~\cite{stretcu2015multiple} \\N}
	    & \shortstack{~\cite{papazoglou2013fast} \\ N}
	    & \shortstack{Ours \\$v1.0$\\ N}
	    & \shortstack{Ours \\$v2.2$\\ N}\\
	    \hline\hline
	    aeroplane 	        & 64.3 	& 75.8    & 51.7    & 38.3    & 65.4     & \textbf{\underline{76.3}} & 76.3\\
	    bird 		        & 63.2 	& 60.8    & 17.5    & 62.5    & 67.3     & \textbf{\underline{71.4}} & 68.5 \\
	    boat                & \underline{73.3} 	& 43.7    & 34.4    & 51.1    & 38.9     & \textbf{65.0} & 54.5\\
	    car                 & 68.9 	& \underline{71.1}    & 34.7    & 54.9    & \textbf{65.2}     & 58.9 & 50.4 \\
	    cat                 & 44.4 	& 46.5    & 22.3    & 64.3    & 46.3     & \textbf{\underline{68.0}} & 59.8\\
	    cow                 & \underline{62.5} 	& 54.6    & 17.9    & 52.9    & 40.2     & \textbf{55.9} & 42.4\\
	    dog                 & \underline{71.4} 	& 55.5    & 13.5    & 44.3    & 65.3     & \textbf{70.6} & 53.5 \\
	    horse               & 52.3 	& \underline{54.9}    & \textbf{48.4}    & 43.8    & \textbf{48.4}     & 33.3 & 30.0\\
	    motorbike           & \underline{78.6} 	& 42.4    & 39.0    & 41.9    & 39.0     & \textbf{69.7} & 53.5 \\
	    train               & 23.1 	& 35.8    & 25.0    & \textbf{\underline{45.8}}    & 25.0     & 42.4 &  60.7\\
	    \hline
	    Avg                 & 60.2 	& 54.1    & 30.4    & 49.9    & 50.1     & \textbf{\underline{61.1}} & 54.9\\
	    \hline\hline
	    \shortstack[l]{time\\ sec/frame} & N/A & N/A & N/A & 6.9 & 4 & \multicolumn{2}{|c|}{\textbf{\underline{0.35}}}  \\
	    \hline
	    \end{tabular}
	    }
	    \end{center}
	    \caption{The CorLoc scores of our method and 5 other state-of-the-art methods, on the YouTube-Objects dataset (note that result for v2.2 of the dataset are not considered for comparison).} \label{tab:YTO}
	    \end{table}

	\subsection{SegTrack v2 dataset}
	    \textbf{Dataset.}
	    The SegTrack dataset was originally introduced by ~\cite{tsai2012motion}, for evaluating tracking algorithms. Further, it was adapted for the task of video object segmentation ~\cite{li2013video}. We work with the second version of the dataset (SegTrack v2), which contains 14 videos ($\approx$ 1000 frames), with pixel level ground truth annotations for the foreground object, in every frame. It contains challenging sequences with complex backgrounds, objects with similar color patterns to the background, a wide variety of object sizes, camera motion and complex object deformations. There are objects from multiple categories: bird, cheetah, human, worm, monkey, dog, frog and parachute. There are 8 videos with annotations for the primary object and 6 videos with 2 or multiple objects. 
	    
	    \textbf{Metric.} For the evaluation on the SegTrack we have adopted the average intersection over union metric. We specify that for the purpose of this evaluation, we use GrabCut for refinement of the soft-segmentation masks.
	    
	    \textbf{Results.} We compare our method against ~\cite{lee2011key, zhang2013video, wang2016video, papazoglou2013fast, li2013video}. We considered their results as originally reported by ~\cite{wang2016video}. The comparison is presented in Table \ref{tab:SegTrack}. As for the YouTube-Objects dataset, we compare our method against both supervised and unsupervised methods, and in the table, we have marked state-of-the-art results for unsupervised methods (bold), and overall state-of-the-art results (underlined). The execution times are also introduced, to highlight that our method outperforms other approaches in terms of speed (see section \ref{subs:compTime}).
	    
	    The performances of our method are competitive, while being an unsupervised method, that benefits of no tuning depending on the testing database.  
	    Also, we prove that our method is one order of magnitude faster than the  previous state-of-the-art, in terms of speed, ~\cite{papazoglou2013fast} (for details see section \ref{subs:compTime}). 
	    	    
	    \begin{table}[h]
	    \begin{center}
	    \resizebox{\columnwidth}{!}{%
	    \begin{tabular}{|l|c|c|c||c|c|c|}
	    \hline 	    
	    \shortstack{Method \\ Supervised?}	    
	    & \shortstack{~\cite{lee2011key} \\ Y} 
	    & \shortstack{~\cite{zhang2013video} \\ Y} 
	    & \shortstack{~\cite{wang2016video} \\ Y}
	    & \shortstack{~\cite{papazoglou2013fast} \\ N}
	    & \shortstack{~\cite{li2013video} \\ N}
	    & \shortstack{Ours \\ N}\\
	    \hline\hline
	    bird of paradise 	& 92 	& -     & \underline{95}    & 66    & \textbf{94}     & 93 \\
	    birdfall 		    & 49 	& \underline{71}    & 70    & 59    & \textbf{63}     & 58 \\
	    frog                & 75 	& 74    & \underline{83}    & \textbf{77}    & 72     & 58 \\
	    girl                & 88 	& 82    & \underline{91}    & 73    & \textbf{89}     & 69 \\
	    monkey              & 79 	& 62    & \underline{90}    & 65    & \textbf{85}     & 69 \\
	    parachute           & \underline{96} 	& 94    & 92    & 91    & 93     & \textbf{94} \\
	    soldier             & 67 	& 60    & \underline{85}    & 69    & \textbf{84}     & 60 \\
	    worm                & \underline{84} 	& 60    & 80    & 74    & 83     & \textbf{\underline{84}} \\
	    \hline
	    Avg                 & 79 	& 72    & \underline{86}    & 72    & \textbf{83}     & 73 \\
	    \hline\hline
	    \shortstack[l]{time\\ sec/frame} & $>$120 &$>$120 & N/A & 4 & 242 & \textbf{\underline{0.35}} \\
	    \hline
	    \end{tabular}
	    }
	    \end{center}
	    \caption{The average IoU scores of our method and 5 other state-of-the-art methods, on the SegTrack v2 dataset.} \label{tab:SegTrack}
	    \end{table}

	\subsection{Computation time} \label{subs:compTime}
	      One of the main advantages of our method is the reduced computational time. Note that all per pixel classifications can be efficiently implemented by linear filtering routines, as all our classifiers are linear. It takes only \textbf{0.35 sec/frame} for generating the soft segmentation masks (initial object cues: 0.05 sec/frame, object proposals refinement: 0.03 sec/frame, patch-based regression model: 0.25 sec/frame, motion estimation: 0.02 sec/frame (Tabel \ref{tab:stagesEval})). The method was implemented in Matlab, with no special optimizations. All timing measurements were performed using a computer with an Intel core i7 2.60GHz CPU. The method of Papazoglou et al. ~\cite{papazoglou2013fast} report a time of 3.5 sec/frame for the initial optical flow computation, on top of which they run their method, which requires 0.5 sec/frame, leading to a total time of 4 sec/frame. The method introduced in ~\cite{stretcu2015multiple} has a total of 6.9 sec/frame. For other methods, like the one introduced in ~\cite{zhang2013video, lee2011key}, it takes up to 120 sec/frame only for generating the initial object proposals using the method of ~\cite{endres2010category}. There is no available information regarding the computational time of other considered methods, but due to their complexity we expect them to be orders of magnitude slower than our method. 

\section{Conclusions}
	We presented an efficient fully unsupervised method for object discovery in video that is both fast and accurate - it achieves state of the art results on a challenging benchmark for bounding box object discovery and very competitive performance on an unsupervised segmentation video dataset. At the same time, our method is fast, being at least an order of magnitude faster than competition. We achieve an excellent combination of speed and performance by exploiting the contrasting properties between objects and their scenes, in terms of appearance and motion, which makes it possible to select positive feature samples with a very high precision. We show theoretically that high precision is sufficient for reliable unsupervised learning (since positives are generally less frequent than negatives), which we perform both at the level of single pixels and at the higher level of groups of pixels to capture higher order statistics about objects appearance, such as local texture and shape. The speed and state of the art accuracy of our algorithm, combined with theoretical guarantees that hold in practice under mild conditions, make our approach unique and valuable in the quest for solving the unsupervised learning problem in video.
    \paragraph{Acknowledgements:} This work was supported by UEFISCDI, under project PN-III-P4-ID-ERC-2016-0007.
    

\end{document}